\documentclass[letterpaper, 10 pt, conference]{ieeeconf}  % Comment this line out if you need a4paper

\makeatletter
\let\NAT@parse\undefined
\makeatother

\IEEEoverridecommandlockouts                              % This command is only needed if 
\usepackage{times}  %Required
\usepackage{helvet}  %Required
\usepackage{courier}  %Required
\usepackage{url}  %Required
\usepackage{graphicx}  %Required
\usepackage{balance}
\usepackage{color}
\usepackage{amsmath}
\usepackage{amssymb}
\usepackage{multirow}
\usepackage[font={small}]{caption}
\usepackage[utf8]{inputenc}
\usepackage{booktabs}
\usepackage{subfig}
\usepackage{dblfloatfix}
\usepackage{tikz}
\usepackage{cite}
\usepackage{algorithm}% http://ctan.org/pkg/algorithms
\usepackage{algpseudocode}% http://ctan.org/pkg/algorithmicx
\usepackage{cancel}
\usetikzlibrary{shapes, arrows}
\usetikzlibrary{backgrounds, fit, calc, 3d}
\usetikzlibrary{positioning}
\definecolor{dred}{rgb}{1,0,0}
\definecolor{dblue}{rgb}{0,0,1}

\renewcommand{\baselinestretch}{0.99}

\usepackage{diagbox}

\newcommand{\eg}{e.\,g.}

\newcommand\figref[1]{Fig.~{\ref{#1}}}
\newcommand\tabref[1]{Table~{\ref{#1}}}

\newcommand\eqnref[1]{{Eq.~\eqref{#1}}}

\newcommand{\Loss}{\mathcal{L}}
\newcommand{\Eb}[2]{\mathbb{E}_{#1}\left[#2\right]}
\newcommand\given[1][]{\:#1\vert\:}

\newcommand{\E}{\mathbb{E}}

\newcommand{\demos}{\mathbf{X}}
\newcommand{\traj}{\mathbf{\Xi}}
\newcommand{\numberofdemos}{j}
\newcommand{\numberofpoints}{n}

\newcommand{\pose}{\mathbf{\xi}}
\newcommand{\dpose}{\dot{\mathbf{\xi}}}
\newcommand{\gmmparam}{\theta}
\newcommand{\deltaparam}{\Delta\gmmparam}
\newcommand{\prior}{\pi}
\newcommand{\mean}{\boldsymbol{\mu}}
\newcommand{\cov}{\boldsymbol{\Sigma}}

\newcommand{\pparam}{\phi}
\newcommand{\qparam}{\psi}
\newcommand{\aeparam}{\omega}

\newcommand{\autoenc}{{AE}_{\aeparam}}
\newcommand{\latent}{\mathbf{z}}
\newcommand{\policy}{\pi_{\pparam}}
\newcommand{\action}{\mathbf{a}}
\newcommand{\state}{\mathbf{s}}
\newcommand{\obs}{\mathbf{o}}
\newcommand{\reward}{r}
\newcommand{\ent}{\mathcal{H}}
\newcommand{\rb}{\mathcal D}
\newcommand{\Q}{Q_{\qparam}}

\newcommand{\ourmodel}{SAC-GMM}
\usepackage[hidelinks]{hyperref}
\usepackage{xcolor}
\hypersetup{
	colorlinks,
	linkcolor={red!50!black},
	citecolor={blue!50!black},
	urlcolor={blue!80!black}
}

\title{\LARGE \bf Robot Skill Adaptation \\ via Soft Actor-Critic Gaussian Mixture Models}
\author{Iman Nematollahi$^\ast$, Erick Rosete-Beas$^\ast$, Adrian Röfer, Tim Welschehold, Abhinav Valada, Wolfram Burgard% <-this % stops a space
\thanks{$^\ast$These authors contributed equally. All  authors  are  with  University of Freiburg, Germany. This work has  been supported by German Federal Ministry of Education and Research under contract number 01IS18040B-OML and the BrainLinks-BrainTools center
of the University of Freiburg.\looseness=-1}
}

\begin{document}

\maketitle
\begin{abstract}
A core challenge for an autonomous agent acting in the real world is to adapt its repertoire of skills to cope with its noisy perception and dynamics. To scale learning of skills to long-horizon tasks, robots should be able to learn and later refine their skills in a structured manner through trajectories rather than making instantaneous decisions individually at each time step. To this end, we propose the Soft Actor-Critic Gaussian Mixture Model (\ourmodel), a novel hybrid approach that learns robot skills through a dynamical system and adapts the learned skills in their own trajectory distribution space through interactions with the environment. Our approach combines classical robotics techniques of learning from demonstration with the deep reinforcement learning framework and exploits their complementary nature. We show that our method utilizes sensors solely available during the execution of preliminarily learned skills to extract relevant features that lead to faster skill refinement. Extensive evaluations in both simulation and real-world environments demonstrate the effectiveness of our method in refining robot skills by leveraging physical interactions, high-dimensional sensory data, and sparse task completion rewards.  Videos, code, and pre-trained models are available at \url{http://sac-gmm.cs.uni-freiburg.de}.
\end{abstract}

\section{Introduction}
Thinking ahead is a hallmark of human intelligence. From early infancy, we form rich primitive object concepts through our physical interactions with the real world and apply this knowledge as an intuitive model of physics for reasoning about physically plausible trajectories and adapting them to suit our purposes~\cite{lake2017building}. This is at odds with most current deep imitation and reinforcement learning paradigms for robot sensorimotor control, which, despite recent progress~\cite{ho2016generative,schulman2017proximal,fujimoto2018addressing,haarnoja2018soft}, are typically trained to make isolated decisions at each time step of the trajectory. In fact, most existing methods for learning manipulation skills are end-to-end high-capacity models that map directly from pixels to actions~\cite{finn2017one,zeng2018robotic,berscheid2020self}. However, although these approaches can capture complex relationships and are flexible to adapt in face of noisy perception, they require extensive amounts of data, and the trained agent is typically bound to take a distinct decision at every time step.\looseness=-1

Learning from demonstration~\cite{billard2008survey} is the classical paradigm to tackle the problem of representing skills with a trajectory-space policy. 
In this context, dynamical systems have shown to be a physically plausible motion generation mechanism that provides a high level of reactivity and robustness against perturbations in the environment~\cite{schaalnonlinear,khansari2011learning,ijspeert2013dynamical,silverio2015learning,manschitz2018mixture}.
%More specifically, the dynamical systems approach for trajectory learning from demonstrations~\cite{schaalnonlinear,khansari2011learning,ijspeert2013dynamical,silverio2015learning,manschitz2018mixture} provides a high level of reactivity and robustness against perturbations in the environment. 
Despite the great success of dynamical systems in affording flexible robotic systems for industry, where a high-precision state of the environment is available, they are still of limited use in more complex real-world robotics scenarios. The main limitations of current dynamical systems in contrast to deep sensorimotor learning methods are their incompetence in handling raw high-dimensional sensory data such as images, and their susceptibility to noise in the perception pipeline.\looseness=-1

\begin{figure}[t]
	\centering
	\includegraphics[width=1\columnwidth]{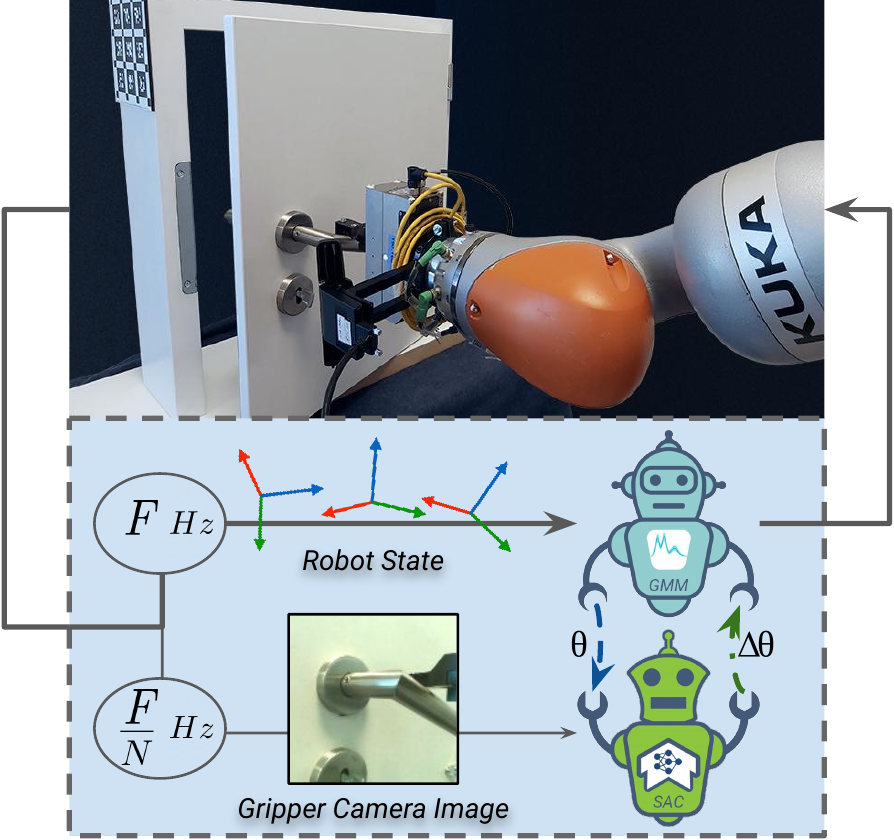}
	\caption{Soft Actor-Critic Gaussian Mixture Models learn and refine real-world robot skills in the trajectory distribution space. While the Gaussian Mixture Model (GMM) agent learns a skill with few demonstrations and controls the robot with high-frequency $F$, the Soft Actor-Critic (SAC) agent affords to work with a lower frequency and refines the skill via adapting GMM parameters by leveraging physical interaction, high-dimensional sensory data, and sparse rewards.}
	\label{fig:introduction}
	\vspace*{-6mm}
\end{figure}

In this paper, we advocate for hybrid models in learning robot skills: ``Soft Actor-Critic Gaussian Mixture Models'' (\ourmodel s, see Fig~\ref{fig:introduction}). \ourmodel s  learn and refine robot skills in the real-world and present a hybrid model that combines dynamical systems and deep reinforcement learning in order to leverage their complementary nature. More precisely, \ourmodel s learn a trajectory-based Gaussian mixture policy of skills from demonstrations and refine it by physical interactions of a soft actor-critic agent with the world. Our hybrid formulation allows the dynamical system to utilize high-dimensional observation spaces and cope with noise in demonstrations and sensory observations while maintaining a reactive and robust trajectory-based policy when interacting with dynamic environments.
We argue that maintaining this physically meaningful structure within the reinforcement learning refinement will yield enhanced performance and stability compared to residual corrections or direct learning of desired end-effector velocities.
The method is simple, sample efficient and readily applicable in a variety of robotics scenarios. We exemplify this, by using our hybrid model for simulated peg insertion and power lever sliding skills, and a real-world door-opening skill. We demonstrate that \ourmodel{} is able to successfully open a door in the real world after half an hour of physical interaction.

The main contributions of this paper are: 
1) a hybrid model for learning and refining skills in trajectory distribution space, 2) exploiting high-dimensional sensory inputs obtainable solely during skill adaptation through physical interaction, such as tactile images, gripper camera images, and static camera depth maps to refine parameters of a dynamical system, 3) mitigating amount of robot exploration efforts for learning skills in sparse reward settings through a dynamical system model learned from few demonstrations, and 4) learning to refine two simulated and one real-world robot manipulation skills.

\section{Related Work}
Learning adaptive skill models has been a long-standing goal in both robotics and machine learning. Our work builds on prior research in this domain which falls into model-based and data-driven paradigms and attempts to bridge the gap between them. 

Learning from demonstration~\cite{billard2008survey} is the paradigm in which the robot learns a new skill from demonstrations presented by humans. Dynamical systems~\cite{khansari2011learning,ijspeert2013dynamical} use these demonstrations to provide an analytical description of robot motion over time. There are different underlying representations of dynamical systems, ranging from movement primitives~\cite{pastor2009learning} to probabilistic approaches~\cite{calinon2010learning}. There exists a large body of recent work proposing augmentations of dynamical systems, \eg, fitting of trajectories in a physically consistent fashion~\cite{figueroa2018physically}, ensuring kinematic feasibility of generated trajectories for a mobile robot~\cite{twelsche18iros}, enabling dynamical systems to generate contact forces~\cite{amanhoud2019dynamical}, integrating non-geometric features into action models~\cite{nematoli19icra}, preventing mode collapse via mixture density networks~\cite{zhou2020movement}, using diffeomorphisms to give rise to inherently stable dynamical system~\cite{rana2020euclideanizing}, and relaxing the point-to-point constraint on demonstrated trajectories~\cite{urain2020imitationflow}. Nonetheless, dynamical systems still suffer from the curse of dimensionality and are ineffective at learning from large or noisy datasets. In contrast, our skill model leverages high-dimensional sensory measurements to adapt its dynamical system and copes with noisy perception and dynamics.

An alternative approach to explicitly modeling robot skills via an analytical framework is to learn an implicit skill model using the robot's interaction data. These data-driven approaches are trained by optimizing directly for skill success. In particular, some recent works have proposed learning robot skills such as grasping~\cite{mousavian20196}, pick-and-stow~\cite{hermann2020adaptive}, and part discovery~\cite{gadre2021act} first in simulation, where interaction is cheap and labeled, and then transferring the agent to the real world. Another transfer learning framework adoption is to learn a vision model from passive observations first and then to leverage the learned representations for learning manipulation skill models~\cite{yen2020learning}. Further approaches attempt to learn the real-world dynamics in 2D~\cite{finn2016unsupervised} or 3D~\cite{nematollahi2020hindsight}, and then use this model to perform specific skills such as pushing~\cite{kloss2020accurate},  peg insertion~\cite{lee2020making}, or ball bearing~\cite{tian2019manipulation}. Recent reinforcement learning (RL) works have proposed learning robot skills using images of the goal~\cite{singh2019end} or learned representations from unlabeled skill-related videos~\cite{sermanet2018time,mees2020adversarial} as the reward function. In general, these high-capacity deep neural networks are not sample-efficient and require a large amount of interaction data to enable a robot to learn complex skills. Moreover, they ignore low-level analytical descriptions of skills and are forced to make individual decisions at each time step. Compared to these approaches, our structured skill model reasons in trajectory space while still optimizing for skill success.\looseness=-1

Our approach for learning adaptive skill models falls under a broader category of hybrid models~\cite{bahl2020neural,zeng2020tossingbot}. These approaches combine classical controllers or dynamical system frameworks with RL and learn robot skills with a fair amount of real-world interaction. More specifically, our hybrid skill model (see \figref{fig:introduction}) leverages both 1) a dynamical system to provide an analytical description of a skill in the trajectory space and 2) a deep reinforcement learning framework to refine the skill via physical interactions and compensate for noisy perception. In contrast to prior work in this domain~\cite{DBLP:journals/corr/abs-1906-05841,davchev2020residual,ranjbar2021residual}, our approach learns to adapt the skill in the trajectory space instead of predicting a residual action at each time step. We demonstrate that our approach mitigates the exploration time needed to refine the robot skill drastically. A further hybrid approach has been introduced by Rey~\emph{et al.}~\cite{Rey:227496}. In contrast to their work, we use a higher dimensional observation for the reinforcement learning refinement of the skill, while their focus is on mathematical properties of the dynamical systems.  \looseness=-1

\section{Problem Formulation}
\begin{figure*}[t]
	\centering
	\includegraphics[scale=0.72]{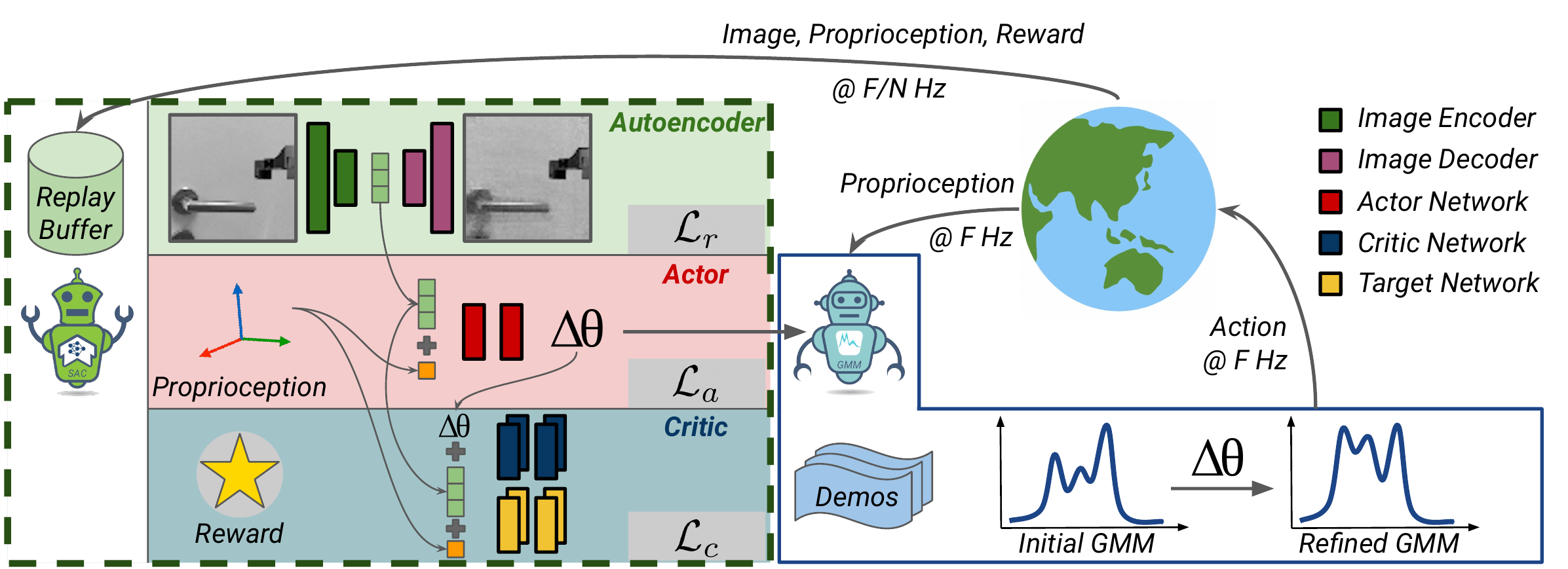}
    \caption{Structure of \ourmodel: Given a set of human demonstrations, the GMM agent learns an initial dynamical system parameterization $\theta$ that provides an analytical description of the robot's skill trajectory. After $N$ interactions by the GMM agent with the environment (working with frequency $F$), the SAC agent receives a high-dimensional observation, robot state, and a sparse success reward from the environment. It then refines the initial GMM agent's trajectory parameters by $\deltaparam$ for the next $N$ interactions to optimize for skill success.}
    \label{fig:architecture}
    \vspace{-0.5cm}
\end{figure*}
We consider a motion to be driven by a dynamic system governed by a set of first order differential equations defining the dependency of the system velocity on the current state. Following this approach, robot skills are directly defined in trajectory space as a global map that specifies the required motion for reaching a target from arbitrary start poses. %that can be encoded using a GMM.
%We aim to learn a hybrid model for robot skills in the trajectory space, \ie, a skill model that learns and refines a global map that instantly specifies the correct direction for reaching the target. 
We are interested in using this structured model to enable a robot to learn and adapt skills quickly in the real world. Our goal is to first learn skills from a few demonstrations and later refine them through the robot's physical interactions in the world.

For this, we consider robot skills as a set of trajectories in 3D space. 
% We denote a robot skill by $\skill = \langle (\pose_1, \obs_1), \dots,(\pose_\numberofpoints, \obs_\numberofpoints) \rangle$, which is a sequence of $\numberofpoints$ vectors $(\pose_i, \obs_i)$, where $\pose_i \in \mathbb{R}^d$ is the robot's geometric pose, and $\obs_i \in \mathbb{R}^{\obsdim}$ is its high-dimensional observation of the environment. 
We denote a trajectory by $\traj = \langle \pose_1, \dots,\pose_\numberofpoints \rangle$, where $\pose_i \in \mathbb{R}^d$ is the robot's geometric pose. Moreover, each execution of a robot skill induces a set of observations $\mathcal{O} = \langle \obs_1, \dots,\obs_\numberofpoints \rangle$ from the environment. Thus, a typical robot skill is described by its 6-DOF (Degree-of-Freedom) trajectory and co-occuring observations such as images, depth maps, or tactile measurements. The observations cannot be controlled directly by the robot but are reactions of the environment to the geometric course of the trajectory. %or consequences of an external source.
Nevertheless, the robot has to adjust its geometric route in response to these observations.\looseness=-1

To learn a robot skill, we require a set of demonstrations $\demos = \langle \traj_1, \dots,\traj_\numberofdemos \rangle$, where each demonstration is a trajectory that exhibits the geometric route of the skill. Note that these demonstrations can be shown either by a human performing the skill or robot teleoperation. Our goal is to first learn a skill model $f_{\theta}(\pose) = \dot{\pose}$ that maps each robot pose to its first derivative (velocity), such that it fulfills the following desiderata: (1) it represents the skill in trajectory distribution space, 
%\ie, it provides a global map that, considering the current pose of the robot and the target, instantly specifies the correct direction for reaching the target, 
and (2) $f$ is a parametric function. The latter ensures that we can later refine our skill model by adapting its parameters $\theta$. Note that this initial learning step does not include the high-dimensional observations as \textit{i)} these might not be easy to capture in demonstrations and \textit{ii)} they cannot be directly included in the parameterization. 

%To refine a robot skill, we require a sparse task completion reward that assesses the robot's performance. 
We then aim to refine the previously learned skill directly in the parameterized trajectory space. The skill refinement policy $\policy(\pose, \obs) = \Delta\theta$, builds on the robot pose and the robot's high-dimensional observations while performing the skill in a sparse task completion reward setup.\looseness=-1

\section{\ourmodel}\label{sec:approach}
We propose the hybrid approach Soft Actor-Critic Gaussian Mixture Models (\ourmodel) consisting of two phases. In the first, we learn a dynamical system parameterization in form of a Gaussian mixture model from a few demonstrations. In the second, we refine this dynamical system with the soft actor-critic algorithm through physical interactions with the world. 
%The key challenges are to learn robot skills from few noisy demonstrations in a trajectory space and to later refine the skills in their own trajectory distribution space in a sparse task completion reward setting. 
The architecture of our hybrid model is shown in Figure~\ref{fig:architecture}.

\subsection{Dynamical System: Gaussian Mixture Model Agent}
Dynamical systems afford an analytical representation of a motion's progression over time, and accordingly, they enable the robot to generate trajectories while being robust in the face of perturbations.% and guarantee convergence to a goal. 
%An autonomous dynamical system is a dynamical system that does not explicitly depend on time and employs a first-order ordinary differential equation to map the robot pose to its velocity. Hence, w
We formulate a robot skill as a control law driven by an autonomous dynamical system, defined by the robot pose $\pose$:
\begin{equation}\label{eq:ODE}
    \dpose = f_{\gmmparam}(\pose) + \boldsymbol{\epsilon},
\end{equation}
where $f_{\gmmparam}$ is the robot skill model, a parametric, non-linear, steady, and continuously differentiable function, and $\boldsymbol{\epsilon}$ is a zero-mean additive Gaussian noise. From a machine learning perspective, learning the noise-free estimate of $f$ from data is a regression problem and can be addressed by a mixture of Gaussians. Given a set of reference demonstrations $\demos$ for a robot skill, we parametrize \eqnref{eq:ODE} through Gaussian Mixture Regression (GMR)~\cite{calinon2010learning}. We first estimate the joint probability density $\mathcal{P}(\dpose, \pose)$ of the robot pose and its corresponding first-order derivative by a Gaussian mixture. Thereby, we parametrize the robot skill model $f$ by $\gmmparam=\{\prior_k,\mean_k, \cov_k \}_{k = 1}^K$, where $\prior_k$ are the priors (or mixing weights), $\mean_k$ the means and $\cov_k$ the covariances of the $k$ Gaussian functions.

%Different methods for fitting $\mathcal{P}(\dpose, \pose)$ to demonstration data exits, the most sophisticated providing stability and convergence guaranties for the dynamical system as Khansari \emph{et al.}~\cite{khansari2011learning} or Figueroa \emph{et al.}~\cite{figueroa2018physically}.
By using this estimated joint probability density function, we employ Gaussian mixture regression (GMR) to retrieve $\dpose$ given $\pose$ as the conditional distribution $\mathcal{P}(\dpose \given \pose)$. 
This way our skill model can reproduce the demonstrated skill by estimating the next velocity at the current robot pose and thus generate a trajectory by updating the pose $\pose$ with the generated velocity $\dpose$ scaled by a time step and proceeding iteratively. For detailed insights on using GMM encoded dynamic system for imitation learning we refer the reader to the extensively available literature~\cite{pastor2009learning,calinon2010learning, ijspeert2013dynamical,khansari2011learning, figueroa2018physically}.
% TODO arxiv version: Refer to appendix.

\subsection{Dynamical System Adaptation: Soft Actor-Critic Agent}
Having learned the skill model $f_{\gmmparam}$, we can now leverage robot interactions with the world to explore and refine the initial model. We formulate this refinement as a reinforcement learning problem in which the agent has to modify the learned skill in the trajectory space and has only access to sparse rewards. In RL, the goal is to learn a policy $\policy$ in a partially observable Markov decision process, consisting of an observation space $\mathcal{O}$, a state space $\mathcal{S}$ and an action space $\mathcal{A}$. In our skill refinement scenario, the agent receives  high-dimensional sensory measurements such as RGB images, tactile measurements, or depth maps which are encoded to a latent representation $\latent$ by an autoencoder.
Together with the robot pose $\pose$, these form our continuous state space. The action space is also continuous and consists of the desired adaptation in the skill trajectory parameters $\deltaparam $. Moreover, the environment emits a sparse reward only if the robot executes the skill effectively. Namely, if $\state_t$, $\action_t$ and $\latent_t$ define the robot's state, action, and latent representation of observation respectively at time step $t$, then 
\begin{equation}
    \state_t := \{\pose_t, \latent_t  \}, \hspace{1em}
    % \quad\text{and}\quad
    \action_t := \{\Delta\prior_k,\Delta\mean_k, \Delta\cov_k \}_{k = 1}^K,
\end{equation}
and consequently
\begin{equation}
    \deltaparam = \policy(\action_t \given \state_t),
\end{equation}
where $\policy$ is the robot skill refinement policy. 
We use $\rho_{\policy}(\state_t,\action_t)$ to denote the state-action marginal of the trajectory distribution induced by the policy $\policy$. 
The robot has to learn this policy from its interactions with the world, such that it maximizes the expected total reward of the refined skill trajectory. Our particular choice for the reinforcement learning framework to learn the skill refinement policy is the soft actor-critic (SAC) algorithm~\cite{haarnoja2018soft}. 
SAC is an off-policy actor-critic method that, in addition to maximizing the expected total reward, aims to maximize the entropy of the stochastic policy. In doing so, SAC encourages robot exploration and avoids converging to non-optimal deterministic policies. Moreover, SAC exhibits good sample efficiency and stability, necessary ingredients for quick skill adaptation in the real world. Therefore, the objective for an optimal skill refinement policy is
\begin{equation}
    \label{eq:maxent_objective}
    J(\policy) = \sum_{t=0}^T\gamma^t \Eb{(\state_t, \action_t)\sim \rho_{\policy}}{ r(\state_t, \action_t) + \alpha \ent(\policy(\cdot | \state_t))},\nonumber
\end{equation}
where $\alpha$ is the temperature parameter that determines the relative importance of the entropy term against the reward and regulates the stochasticity of the policy. 
Our SAC agent stores a collection of $\{\state_t, \obs_t, \action_t, \reward_t, \state_{t+1}, \obs_{t+1}\}_{i=1}^T$ transition tuples in a replay buffer $\rb$, and concurrently learns an autoencoder $\autoenc$, a policy $\policy$ and two Q-functions ${\Q}_1$ and ${\Q}_2$ (to prevent overly optimistic value estimates) and their target networks. More concretely, we use the autoencoder $\autoenc$ to learn a low-dimensional latent representation of the robot's high-dimensional observations. We adopt a similar strategy to the Contractive Autoencoders~\cite{rifai2011contractive} and enforce an L2 penalty on the learned representation to encourage robust features. For training the autoencoder, actor, and critic networks, SAC randomly samples a batch from the replay buffer and performs stochastic gradient descent on minimizing the following loss objectives for autoencoder, critic, and actor respectively:
\begin{align*}
    \label{eq:rae_loss}
    \Loss_{r}(\aeparam, \rb)&= \E_{\obs_t \sim \rb} \big[ \log p_{\aeparam}(\obs_t|\latent_t) + \lambda_{\latent} ||\latent_t||^2 \big],
    \\
    \Loss_{c}(\qparam_i, \rb) &= \E_{(\state_t, \action_t) \sim \rb}\big[{(Q_{\qparam_i}(\state_t, \action_t) - Q^{\text{tar}}(\state_t, \action_t))^2}\big], 
    \\
    \Loss_{a}(\pparam, \rb) &= \E_{\substack{\state_{t} \sim \rb \\ 
                      \action_{t} \sim \policy}} \big[{\alpha\log\policy(\action_t|\state_t) \!- \!\min_{i=1,2}\!{\Q}_{i}(\state_t, \action_t)}\big],
\end{align*}
where $Q^{\text{tar}}$ is the target for the $Q$ functions and is computed using the immediate reward, the value estimate of target $Q$ network and an entropy regularization term.

% \begin{equation}
% \label{eq:sac-q-obj}
%     \begin{split}
%       \Loss_{critic}(\qparam_i, \rb) \!=\!\E_{(\state_t, \action_t) \sim \rb}\big[{(Q_{\qparam_i}(\state_t, \action_t) - Q^{\text{tar}}(\state_t, \action_t))^2}\big], 
%     % \\ \mathrm{where}\ 
%     %   Q^{\text{tar}}(\state_t, \action_t) = \reward(\state_t, \action_t) + 
%     %   \\ \gamma \E_{\substack{\state_{t+1} \sim p \\ 
%     %                       \action_{t+1} \sim \policy}} [\min_{j=1,2}\hat{Q}_{j}(\state_{t+1}, \action_{t+1}) 
%     %   - \alpha\log(\policy(\action_{t+1}| \state_{t+1}))]
%     \end{split}
% \end{equation}
% \begin{equation}
%     \label{eq:sac-pi-obj}
%     \Loss_{actor}(\pparam, \rb) \!= \!\E_{\substack{\state_{t} \sim \rb \\ 
%                           \action_{t} \sim \policy}} \big[{\alpha\log\policy(\action_t|\state_t) \!- \!\min_{i=1,2}\!{\Q}_{i}(\state_t, \action_t)}\big]
% \end{equation}

\subsection{Full Model}
Figure~\ref{fig:architecture} shows how our hybrid model learns and refines a robot skill. The GMM agent is fitted on the provided demonstrations and represents a dynamical system, controlling the motion in the trajectory space. After each $N$ interaction steps with the world driven by the GMM encoded dynamics, the SAC agent receives the current state $\state_t$ consisting of the latent observation $\latent_t$ and the robot state $\pose_t$, and additionally a reward $\reward_t$ for the previous step. It then generates an action $\action_t := \{\Delta\prior_k,\Delta\mean_k, \Delta\cov_k \}_{k = 1}^K$ according to the current $\state_t$. Note that with the latent observation  $\latent_t$ the state $\state_t$ contains more information than the data the original GMM agent was trained on. The SAC agent's action output $\{\Delta\prior_k,\Delta\mean_k, \Delta\cov_k \}_{k = 1}^K$ is then used to adapt the original GMM for the next $N$ interactions with the environment.
The SAC agent optimizes $\policy(\action_t \given \state_t)$ for skill success using ADAM to optimize the autoencoder, critics, and actor networks. The learning rate for the autoencoder is $3\times10^{-5}$, and for critics and actor is $3\times10^{-4}$. The observations that the autoencoder receives are $64\times64$ pixels. In all our experiments, we use $N=32$ and $K=3$.
While updating GMM parameters, we ensure that mixing weights sum to one and covariance matrices stay positive semi-definite~\cite{higham1988computing}. 
At inference time, the procedure remains the same. The robot interacts with the environment based on the GMM dynamic system encoding, while after each $N$ steps, the SAC agent receives information about the current state and skill progress and adapts the original GMM agent accordingly.\looseness=-1

\section{Experimental Evaluation}\label{sec:experiments}
We evaluate \ourmodel{} for learning robot skills in both simulated and real-world environments.  The goals of these experiments are to investigate: (i) whether our hybrid model is effective in performing skills in realistic noisy environments, (ii) if exploiting high-dimensional data boosts the dynamical system adaptation, and (iii) how refining robot skills in trajectory space compares with alternative exploration policies in terms of accuracy and exploration budget.

\subsection{Experimental Setup}
\begin{figure}[t]
\centering
\includegraphics[height=3.5cm]{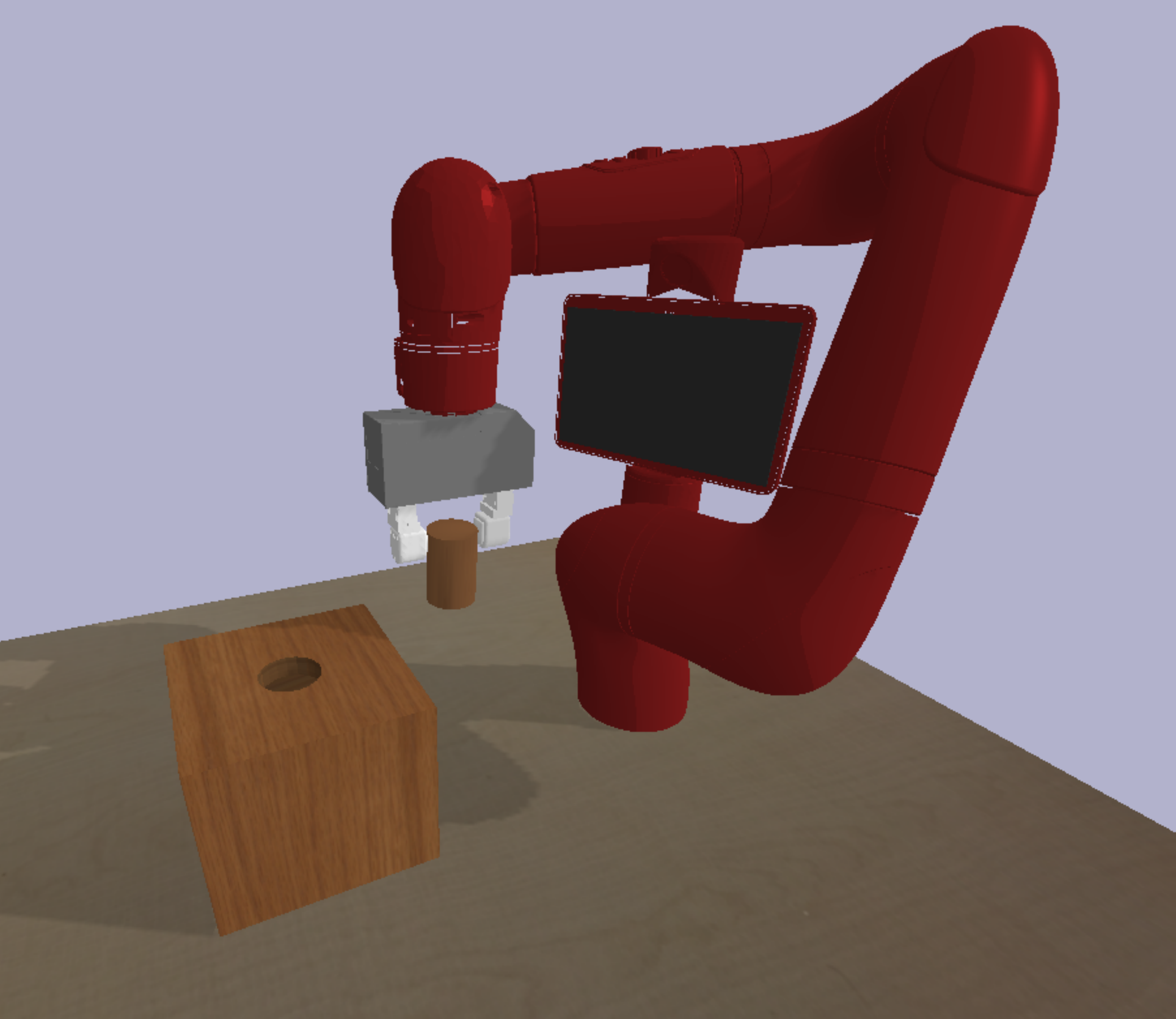}
\includegraphics[height=3.5cm]{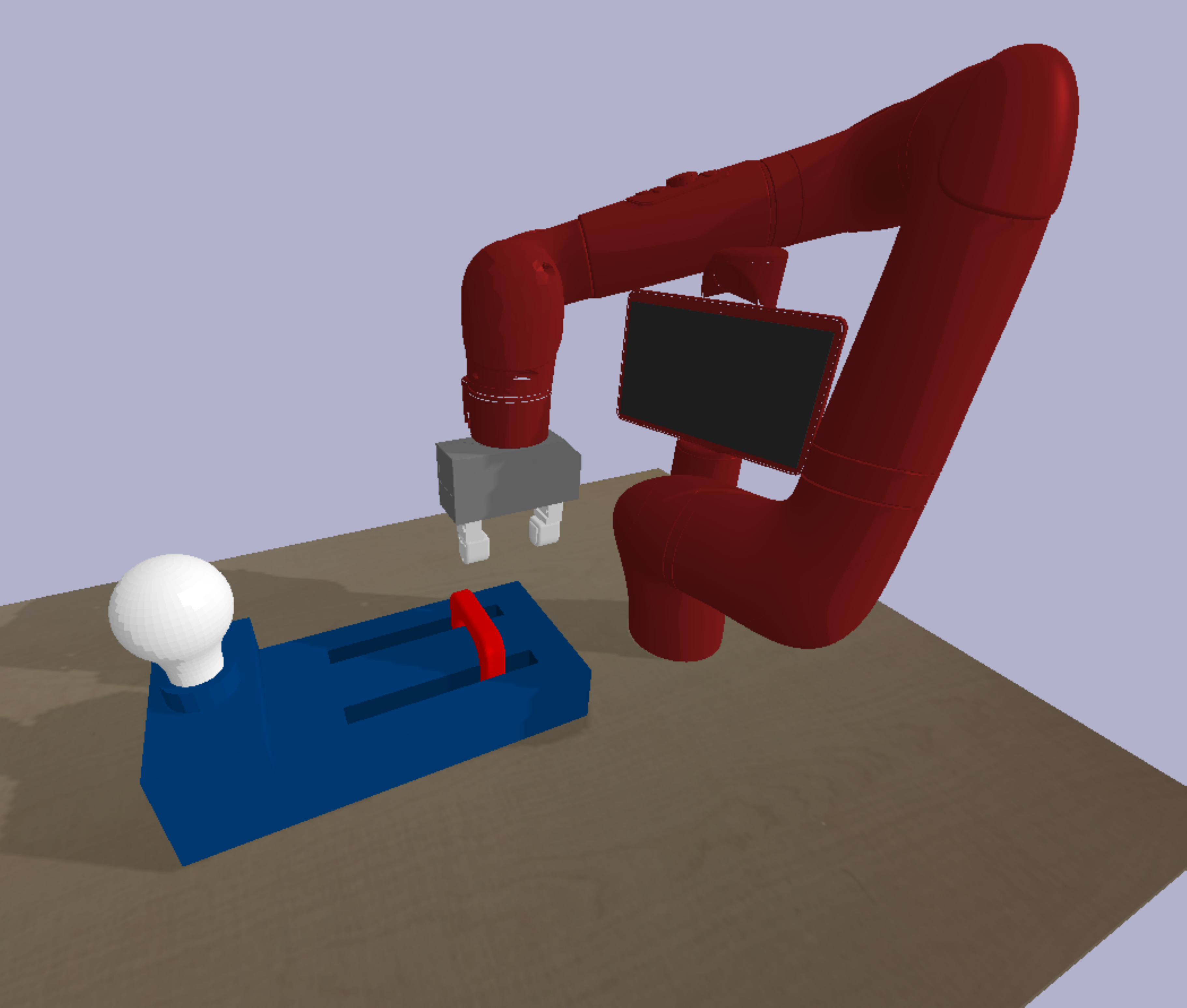}
\caption{Left: Simulated peg insertion scenario; Right: Simulated power lever sliding scenario }
\label{fig:sim_scenarios}
\vspace{-0.7cm}
\end{figure}

We evaluate our approach in both simulated and real-world environments. We investigate two robot skills in simulation: Peg Insertion, and Power Lever Sliding. The environments are simulated with PyBullet and are shown in Figure~\ref{fig:sim_scenarios}. For the peg insertion skill, we design a cylindrical peg and its corresponding circular hole with 50 mm and 56 mm diameters, respectively. The robot receives a success reward whenever the peg is inserted correctly in the hole. We recognize the importance of the sense of touch in fitting a peg smoothly in a hole. Hence, during the skill adaptation phase, we provide the robot with vision-based tactile sensors~\cite{wang2020tacto,lambeta2020digit} on both fingers. For the power lever sliding skill, we design a lever with a relative height of 50 mm to the base. The robot is rewarded whenever it grasps the lever accurately and slides it to the end, such that the light is turned on. For this skill, we provide the robot with static depth perception during the skill refinement to facilitate accurate lever pose estimation. In each demonstration as well as training and test episodes, we randomly change the relative position of the robot to the hole and lever. Moreover, to simulate realistic real-world conditions, we also consider noise regarding the detected position of the hole and lever. Namely, in both environments, we add Gaussian noise with a standard deviation of 1 cm in the target pose's x, y, and z dimensions. For both skills, we collect 20 demonstrations by teleoperating the robot in the simulation. 

\begin{figure}[t]
    \centering
    \includegraphics[scale=0.25]{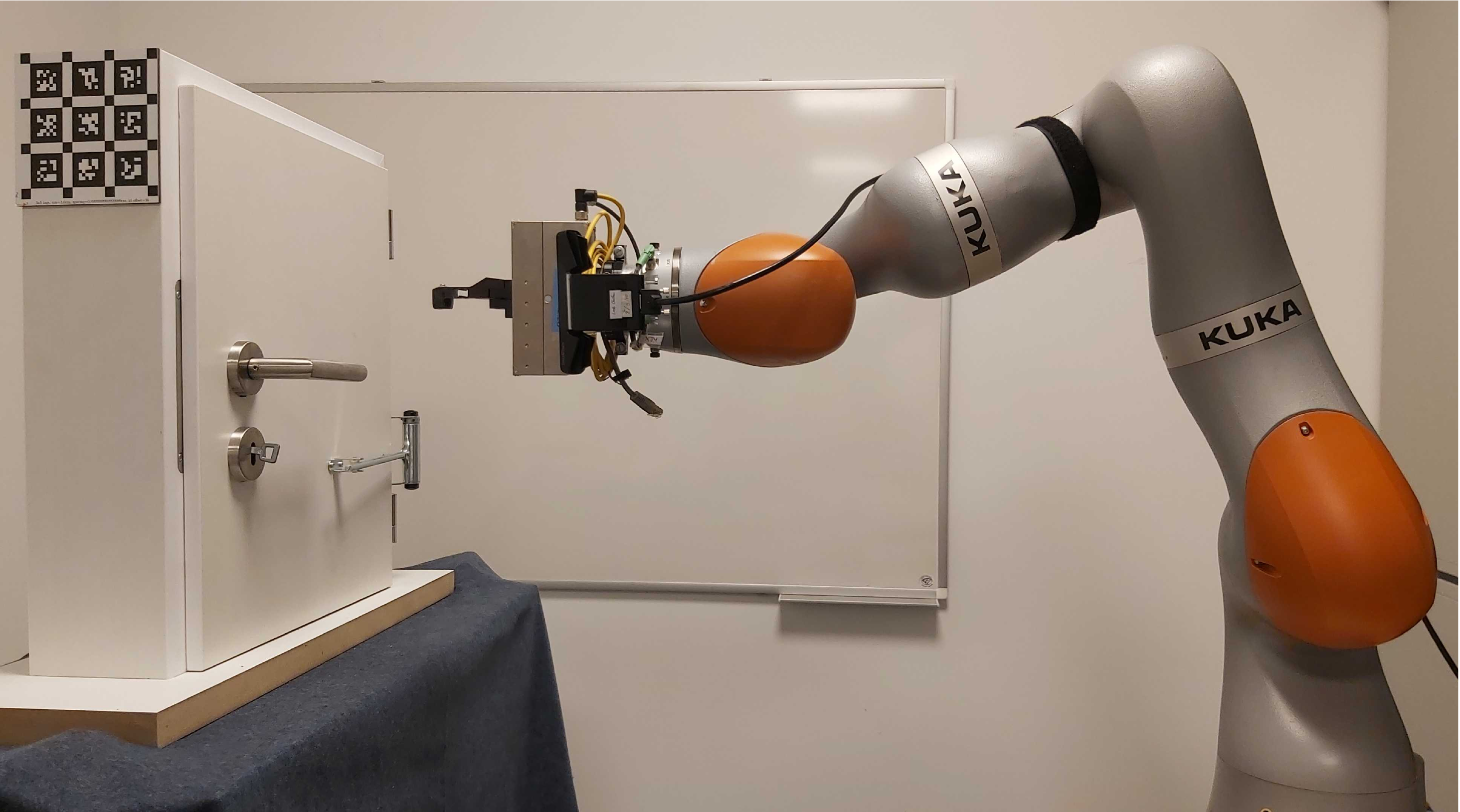}
    \caption{Real-world door opening. While skill refinement, the robot gets a reward when it opens the door. Wrist-mounted camera provides high-dimensional observations during task execution.}
    \label{fig:real_robot_setup}
    \vspace{-0.5cm}
\end{figure}

For the real-world experiment, we investigate a door opening skill, and place our KUKA iiwa manipulator in front of a miniature door, see Figure~\ref{fig:real_robot_setup}. We attach an ArUco marker on the backside of the door to detect when it was successfully opened and grant a reward accordingly. Moreover, to enable our robot to run autonomously without any human intervention, we equip our door with a door closing mechanism. Thus, the door shuts when the robot releases the door handle and starts a new interaction episode. We provide our robot with an Intel SR300 camera mounted on the gripper for an RGB eye-in-hand view during the skill adaptation phase. We use OpenPose~\cite{cao2019openpose} to track the human hand and collect 5 human demonstrations of the door opening skill. Naturally, due to the difference in it's geometry, it is not ideal for the robot to exactly reproduce the observed human hand poses and a task dependent offset to correct this difference would be required for ideal performance of the GMM. However, our approach can also cope with this additional source of noise. 
For the real-world experiments the initial GMM model is fitted using LPV-DS~\cite{figueroa2018physically} while in simulation we rely on SEDS~\cite{khansari2011learning}.
While these dynamic system encodings guarantee stability and convergence, our performed modification does not mathematically maintain these guarantees. However, since the adaptions are small, we did not face any issues with diverging or unstable trajectories in unseen starting configurations. All reported failure cases stem from lacking precision of the motions.\looseness=-1

\vspace{-0.2cm}

\subsection{Evaluation Protocol}
We compare our skill model against the following models:

\noindent\textbf{GMM}: This baseline corresponds to the same dynamical system that we learn with the provided demonstrations in the first step of our approach. 
%We use SEDS~\cite{khansari2011learning} in the simulation experiments framework while for the real world experiments we rely on LPV-DS~\cite{figueroa2018physically} to fit our GMM agents to the demonstrated trajectories. 
However, it is not able to explore or leverage high-dimensional observations to refine its performance.

\noindent\textbf{SAC}: We employ the soft actor-critic agent~\cite{haarnoja2018soft} to explore and learn the skills. We initialize the replay buffer of the SAC agent with the demonstrations of skills. This baseline employs the same SAC structure that we use in \ourmodel{}, including the autoencoder for high-dimensional observations and the network architecture. However, this baseline does not have access to a dynamical system, and consequently, does not reason on the trajectory level.

\noindent\textbf{Res-GMM}: Analogous to our approach, this baseline first learns a GMM agent using the demonstrations and then employs a SAC agent for skill refinement. In contrast to our approach,this SAC agent acts at each time step (instead of each N step), and instead of predicting change in trajectory parameters, it predicts a residual velocity which is summed up with the GMM agent's predicted velocity. This baseline is inspired by recent approaches in the Residual RL domain~\cite{DBLP:journals/corr/abs-1906-05841,davchev2020residual,ranjbar2021residual}. We do not replicate these approaches directly as we want to evaluate the policy refinement and not the choice of underlying dynamical system in which these recent approaches differ.

For the quantitative evaluation of skill models, we employ them to perform the skill 10 times at each evaluation step. We report the average accuracy of each skill model over these episodes in our plots. 
\vspace{-0.1cm}

\subsection{Experiments in Simulation}
\begin{table}[b]
\vspace{-0.5cm}
\centering
\setlength\tabcolsep{1.8pt}
\begin{tabular}{ |c|c|c| } 
\hline
\diagbox{Model}{Task} & Peg Insertion & Lever Sliding \\
\hline
% GMM  & 20.00 & 54.00 \\ 
% SAC  & 0.00 & 0.00 \\ 
% Res-GMM  & 52.88 & 69.86\\
% \ourmodel{} & $\mathbf{92.71}$ & $\mathbf{83.42}$ \\ 
GMM  & 20\% & 54\% \\ 
SAC  & 0\% & 0\% \\ 
Res-GMM  & 30\% & 62\%\\
\textbf{\ourmodel{}} & $\mathbf{86\%}$ & $\mathbf{81\%}$ \\ 
\hline
\end{tabular}
\caption{Our \ourmodel{} outperforms baseline skill models significantly in both robot skills by refining the initial dynamical systems (GMM) substantially.}
\label{table:main_result}
\end{table}

\begin{figure*}[t]
    \centering
    \subfloat[Simulated Peg Insertion]{
        \includegraphics[width=0.33\textwidth]{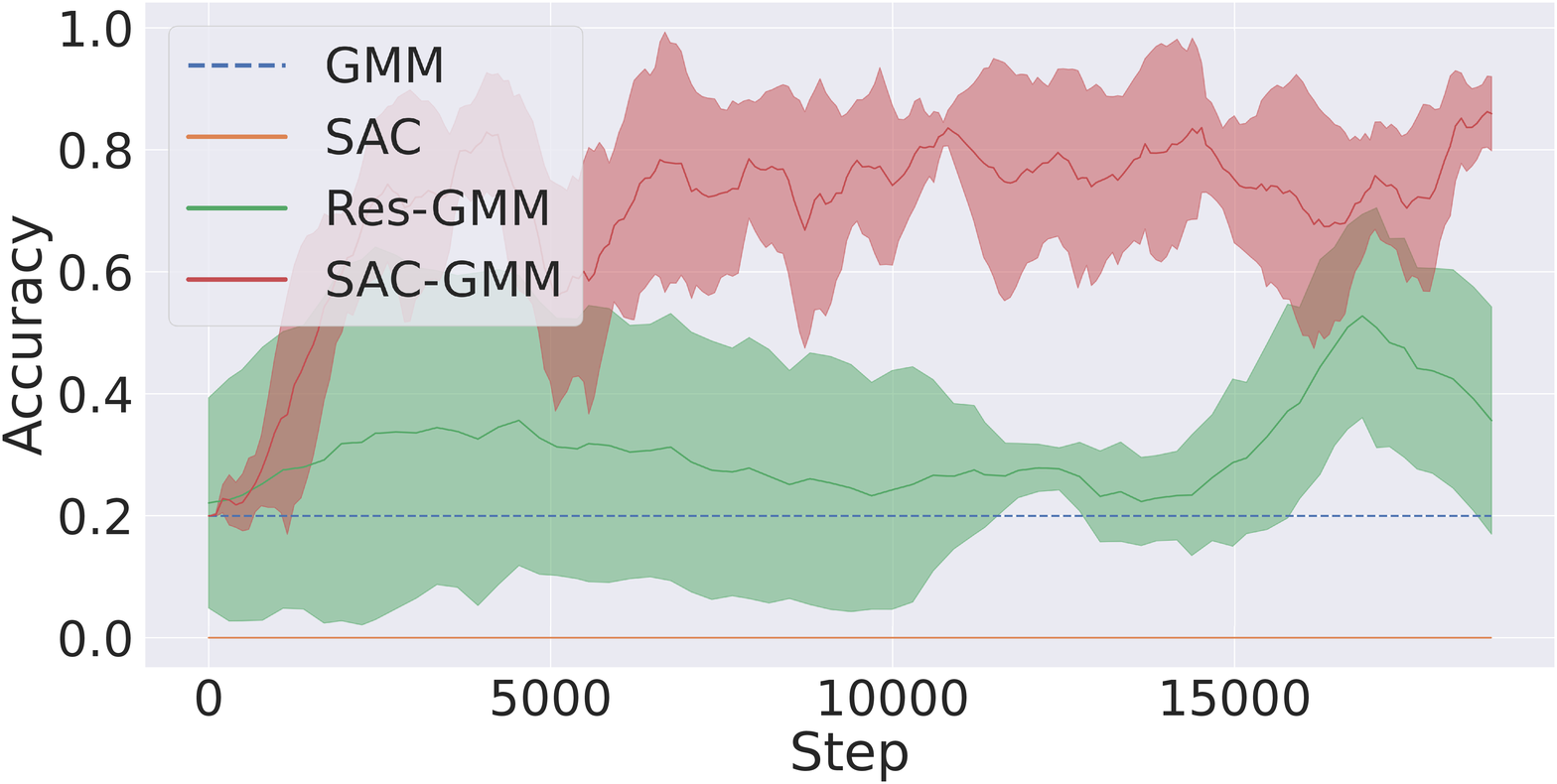}\label{fig:peg_training}}
    \subfloat[\ourmodel{} in noisy Peg Insertion]{
        \includegraphics[width=0.33\textwidth]{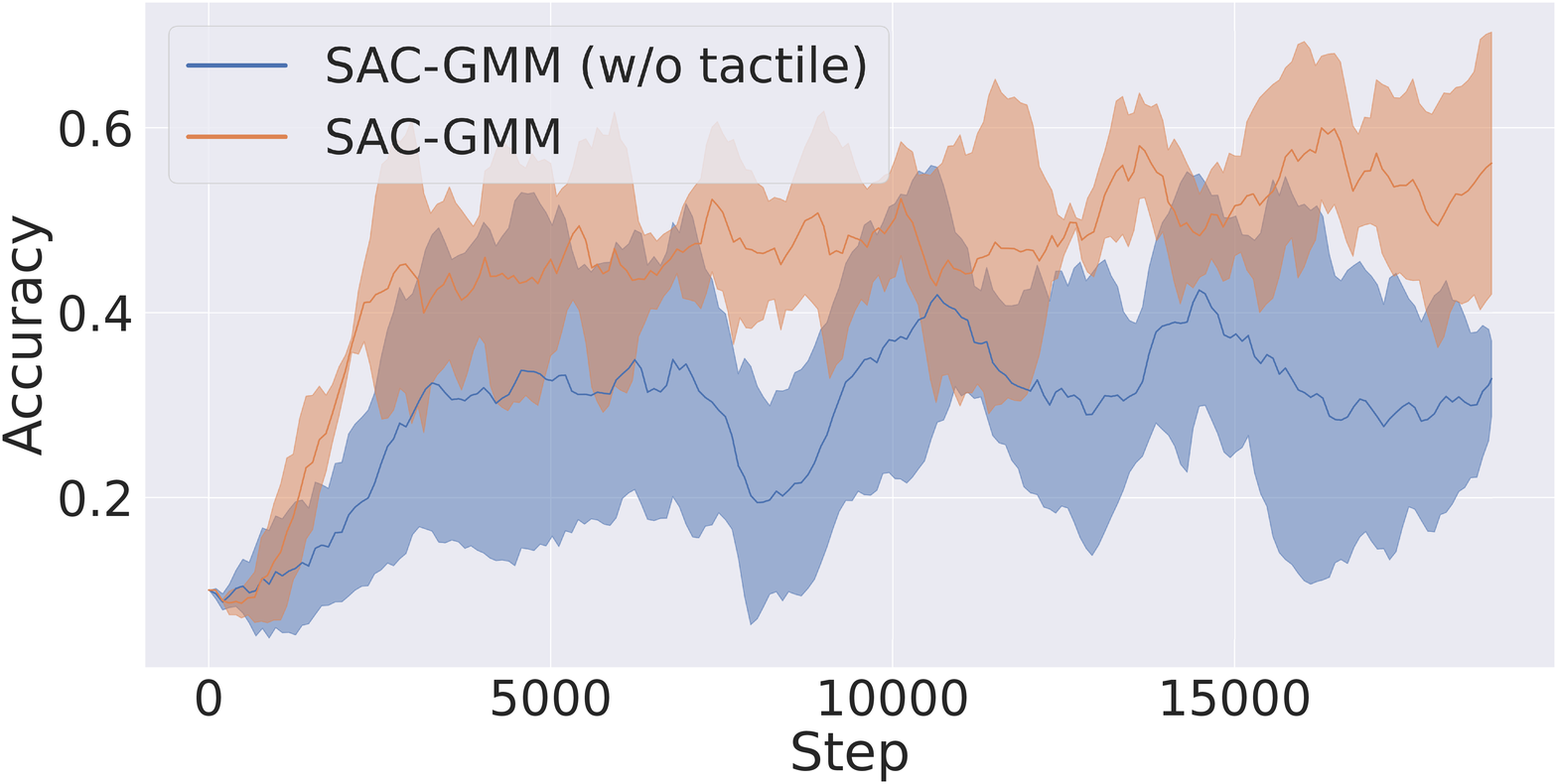}\label{fig:tactile_noise}}
    \subfloat[Real-World Door Opening]{
        \includegraphics[width=0.33\textwidth]{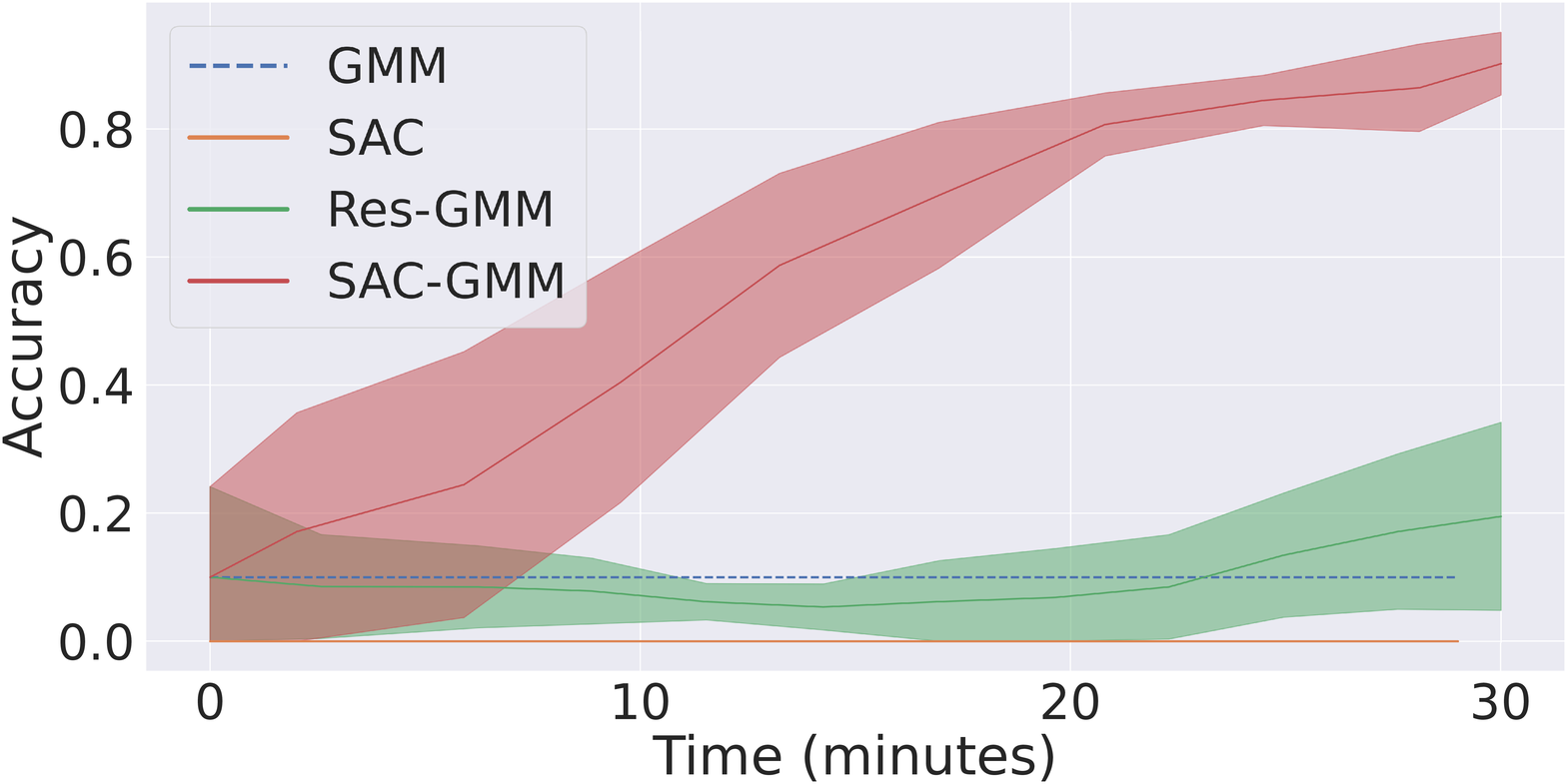}\label{fig:real_world_experiment}}
  	\caption{(a) \ourmodel{}  vs. baseline models for the peg insertion skill, (b) \ourmodel{} exploits high-dimensional data (tactile measurements here) to cope better with noisy perception (c) \ourmodel{} learns to open a real door after half an hour of physical interactions.}
	\label{fig:plots}
    \vspace{-0.6cm}
\end{figure*}

We start by evaluating our method on the peg insertion and power lever sliding skills in simulation. Quantitative results of the average success rate of each skill model over 100 trials per five different random seeds are reported in Table~\ref{table:main_result}.  
Our \ourmodel{} successfully performs the peg insertion and lever sliding skills with an average final success rate of $86\%$ and $81\%$ respectively. It achieves significantly higher success rates than the GMM baseline, proving its effectiveness in refining the robot skills through physical interaction. \figref{fig:peg_training} shows the success rate of all skill models during training for the peg insertion skill. The SAC baseline which uses skill demonstrations as initialization for its replay buffer, performs poorly and cannot learn the robot skills in the sparse reward setting. Though the Res-GMM skill model which adds a residual velocity to the GMM at each time step achieves a notable success rate of $30\%$, it is much slower in skill adaptation than  \ourmodel{} and suffers from a higher variance. This is caused by Res-GMM's inability to reason about skill refinement on the trajectory level and its obligation to take a residual corrective action at each time step of the trajectory.\looseness=-1

To analyze the influence of noise in the perception pipeline and the efficacy of \ourmodel{} in exploiting high-dimensional sensory data, we conduct several experiments on the simulated environments (see \tabref{table:ablation}). We observe that \ourmodel{} can fully leverage high-dimensional data such as tactile measurements and depth maps in all scenarios to achieve a superior skill success rate. \figref{fig:tactile_noise} shows the success rate of \ourmodel{} with and without having access to the tactile sensors~\cite{wang2020tacto} during refinement of the peg insertion skill in the noisy setup. We find that \ourmodel{} utilizes the high-dimensional observation to deal better with noise and learn the skill faster.
\begin{table}[b]
\vspace{-0.5cm}
\centering
\setlength\tabcolsep{1.8pt}
\begin{tabular}{ |c|c|c|c|c|c|c|c|c| } 
\hline
\multirow{5}{*}{\diagbox{Model}{Task}} & \multicolumn{4}{c|}{Peg Insertion}  & \multicolumn{4}{c|}{Lever Sliding}\\
\cline{2-9}
& \multicolumn{2}{c|}{No } & \multicolumn{2}{c|}{With } & \multicolumn{2}{c|}{No } & \multicolumn{2}{c|}{With } \\ 
& \multicolumn{2}{c|}{Noise} & \multicolumn{2}{c|}{Noise} & \multicolumn{2}{c|}{Noise} & \multicolumn{2}{c|}{Noise} \\ 
\cline{2-9}
& No & With & No & With & No & With & No & With  \\ 
& Tactile & Tactile & Tactile & Tactile & Depth & Depth & Depth & Depth  \\ 
\hline
GMM  & 20\% & x & 10\% & x & 54\% & x & 39\% & x \\ 
SAC  & 0\% & 0\% & 0\% & 0\% & 0\% & 0\% & 0\% & 0\% \\ 
Res-GMM  & 24\% & 30\% & 26\% & 23\% & 49\% & 62\% & 29\% & 38\%\\
\textbf{\ourmodel{}}  & $\mathbf{44\%}$ & $\mathbf{86\%}$ & $\mathbf{33}\%$ & $\mathbf{56}\%$ & $\mathbf{68}\%$ & $\mathbf{81}\%$ & $\mathbf{42}\%$ & $\mathbf{52}\%$\\ 
\hline
\end{tabular}
\caption{The average success rate of skill models over 100 trials per five different random seeds, under various noise and sensors settings for the simulated robot skills.}
\label{table:ablation}
\end{table}
We observe that the pure SAC model which does not utilize dynamical systems and accordingly does not reason in the trajectory space cannot learn robot skills in sparse reward settings. We further analyze the SAC skill model in the shaped reward settings. The results in \tabref{table:reward_shape} show how \ourmodel{} stands out in terms of performance and exploration budget. \ourmodel{} outperforms SAC even though it only requires sparse rewards and a third of exploratory interaction steps in the world.
\begin{table}[t]
\vspace{0.1cm}
\centering
\begin{tabular}{ |c|c|c|c|c|c } 
\hline
Model & Reward & Interactions & Peg Insertion & Lever Sliding\\
\hline
SAC & Dense & $60$K & 60\% & 30\%\\ 
\textbf{\ourmodel{}}  & Sparse & $20$K & $\mathbf{86\%}$ & $\mathbf{81\%}$\\ 
\hline
\end{tabular}
\caption{\ourmodel{} significantly outperforms SAC, even though SAC has access to shaped rewards and takes three times more exploratory interaction steps in the world.}
\label{table:reward_shape}
\vspace{-0.6cm}
\end{table}

\vspace{-0.1cm}

\subsection{Real-World Door Opening}
\figref{fig:real_world_experiment} reports the results for the door opening skill in the real world. We find that, although the initial dynamical system (the GMM agent fitted on human demonstrations) enables the robot to reach the door handle, the robot misses the proper position to apply its force and can only open the door with a $10\%$ success rate. This failure is due to the robot's noisy perception and dynamics. Our \ourmodel{} exploits the wrist-mounted camera RGB images and sparse door opening rewards and achieves a $90\%$ success rate after only half an hour of physical interactions ($\sim$100 episodes) with the door. The SAC baseline fails to learn the skill and the Res-GMM model performs poorly, as adding residual velocities at each time step results in non-smooth trajectories. Videos of these experiments are available at \url{http://sac-gmm.cs.uni-freiburg.de}.
\vspace{-0.1cm}

\section{Conclusions}
In this work, we present “Soft Actor-Critic Gaussian Mixture Models” as a new framework for learning robot skills. This hybrid model leverages reinforcement learning to refine robot skills represented via dynamical systems in their trajectory distribution space and exploits the natural synergy between data-driven and analytical frameworks. Extensive experiments carried out in both simulation and real-world settings, demonstrate that our proposed skill model: 1) learns to refine robot skills through physical interactions in realistic noisy environments, 2) exploits high-dimensional sensory inputs available during skill refinement to cope better with noise, and 3) performs robot skills significantly better than comparable alternatives considering the performance accuracy and exploration costs. 

An interesting extension of our work is to investigate transferring skills learned in one environment to new environments, where the visual appearance of the overall setting will differ considerably while the goals of the skills remain similar (\eg, transfer the skill of door opening in one room to other rooms), and refining skills in new environments.
While for the initial dynamic system stability and convergence guarantees can be formulated, the modifications on the model performed online by our SAC agent do not formally abide by these guarantees. Even though we did not experience any issues with divergence or instabilities, further adjustments as suggested by Rey \emph{et al.}~\cite{Rey:227496} could be included to guarantee convergence and stability.  
%\section*{Acknowledgements}
%BLBT
% This work was partly funded by the European Union’s Horizon 2020 research and innovation program under grant
% agreement No 871449-OpenDR and a research grant from Eva Mayr-Stihl Stiftung.

%%%%%%%%%%%%%%%%%%%%%%%%%%%%%%%%%%%% BIBLIOGRAPHY %%%%%%%%%%%%%%%%%%%%%%%%%%%%%%%%%%%%%%%%%%%%

\footnotesize
\balance
\bibliographystyle{IEEEtran}
\bibliography{icra22}

% \newpage
% \section{Appendix}
% \input{7_Appendix}
%%%%%%%%%%%%%%%%%%%%%%%%%%%%%%%%%%%% BIBLIOGRAPHY %%%%%%%%%%%%%%%%%%%%%%%%%%%%%%%%%%%%%%%%%%%%

%\addtolength{\textheight}{-12cm}   % This command serves to balance the column lengths on the last page of the document manually. It shortens the textheight of the last page by a suitable amount. This command does not take effect until the next page so it should come on the page before the last. Make sure that you do not shorten the textheight too much.
\end{document}